\documentclass[twocolumn, switch]{article} 

\usepackage{preprint}

\usepackage{enumitem}
\usepackage{amsmath, amsthm, amssymb, amsfonts}
\usepackage{bm}
\usepackage{amsmath}
\usepackage{graphicx}
\usepackage{subfigure}
\usepackage[algoruled,algo2e]{algorithm2e}
\usepackage{tcolorbox}
\usepackage{setspace}
\usepackage{amssymb}
\usepackage{booktabs}
\usepackage{array}
\usepackage{color}
\usepackage{multirow}
\usepackage{multicol}
\usepackage{threeparttable}
\usepackage{url}
\usepackage[page]{appendix}

\usepackage[numbers,square]{natbib}
\usepackage[utf8]{inputenc}	
\usepackage[T1]{fontenc}	
\usepackage{xcolor}		
\usepackage[colorlinks = true,
            linkcolor = blue,
            urlcolor  = blue,
            citecolor = blue,
            anchorcolor = black]{hyperref}	
\usepackage{booktabs} 		
\usepackage{nicefrac}		
\usepackage{microtype}		
\usepackage{lineno}		
\usepackage{float}			

\usepackage{newfloat}
\DeclareFloatingEnvironment[name={Supplementary Figure}]{suppfigure}
\usepackage{sidecap}
\sidecaptionvpos{figure}{c}

\usepackage{titlesec}
\titlespacing\section{0pt}{12pt plus 3pt minus 3pt}{1pt plus 1pt minus 1pt}
\titlespacing\subsection{0pt}{10pt plus 3pt minus 3pt}{1pt plus 1pt minus 1pt}
\titlespacing\subsubsection{0pt}{8pt plus 3pt minus 3pt}{1pt plus 1pt minus 1pt}

\usepackage{graphics}
\usepackage{hyperref}
\usepackage{color}
\usepackage{multirow}
\usepackage{hhline}
\usepackage{subfigure}
\usepackage{lipsum}

\title{LongCaptioning: Unlocking the Power of Long Video Caption Generation in Large Multimodal Models}

\usepackage{authblk}

\author[1]{Hongchen Wei}
\author[1]{Zhihong Tan}
\author[2]{Yaosi Hu}
\author[2]{Chang Wen Chen}
\author[1]{Zhenzhong Chen*}
\affil[1]{School of Remote Sensing and Information Engineering, Wuhan University}
\affil[2]{Department of Computing, Hong Kong Polytechnic University}

\begin{document}

\twocolumn[ 
  \begin{@twocolumnfalse} 
  
\maketitle

\begin{abstract}

  Large Multimodal Models (LMMs) have demonstrated exceptional performance in video captioning tasks, particularly for short videos. 
	However, as the length of the video increases, generating long, detailed captions becomes a significant challenge. 
	In this paper, we investigate the limitations of LMMs in generating long captions for long videos. Our analysis reveals that open-source LMMs struggle to consistently produce outputs exceeding 300 words, leading to incomplete or overly concise descriptions of the visual content. 
	This limitation hinders the ability of LMMs to provide comprehensive and detailed captions for long videos, ultimately missing important visual information. 
	Through controlled experiments, we find that the scarcity of paired examples with long-captions during training is the primary factor limiting the model's output length. 
	However, manually annotating long-caption examples for long-form videos is time-consuming and expensive. 
	To overcome the annotation bottleneck, we propose the \textbf{LongCaption-Agent}, a framework that synthesizes long caption data by hierarchical semantic aggregation. 
	Using LongCaption-Agent, we curated a new long-caption dataset, \textbf{LongCaption-10K}. 
	We also develop \textbf{LongCaption-Bench}, a benchmark designed to comprehensively evaluate the quality of long captions generated by LMMs. 
	By incorporating LongCaption-10K into training, we enable LMMs to generate captions exceeding 1,000 words for long-form videos, while maintaining high output quality. 
	In LongCaption-Bench, our model achieved State-of-The-Art performance, even surpassing larger proprietary models like GPT4o. 

\end{abstract}

\vspace{0.4cm}

  \end{@twocolumnfalse} 
] 

\newcommand\blfootnote[1]{%
\begingroup
\renewcommand\thefootnote{}\footnote{#1}%
\addtocounter{footnote}{-1}%
\endgroup
}

\section{INTRODUCTION}
\label{sec:intro}
{\blfootnote{Corresponding author: Zhenzhong Chen, E-mail:zzchen@ieee.org}}

Video captioning, a critical task in multimodal understanding, requires models to generate comprehensive and context-rich descriptions for video content. 
With the rapid development of Large Multimodal Models (LMMs)~\cite{zhang2024long,xue2024longvila,liu2024kangaroo,ataallah2024minigpt4,li2023llamavid,jin2023chat-univi,song2024moviechat,xu2024pllava,lin2023video}, video captioning has achieved significant progress. 
It generates descriptions for videos by concatenating visual features and textual features as input to large language models. 
However, previous research has predominantly focused on short videos, where the scope of context is limited, and captions typically remain concise. 
In contrast, long videos introduce additional challenges: they contain more complex, diverse content that requires models to capture and maintain long-term dependencies across frames and segments. 
This increased temporal span makes it more difficult for models to generate captions that are not only comprehensive but also coherent and contextually consistent over extended periods.

\begin{figure*}[t]
	\centering
	\includegraphics[width=0.99\linewidth]{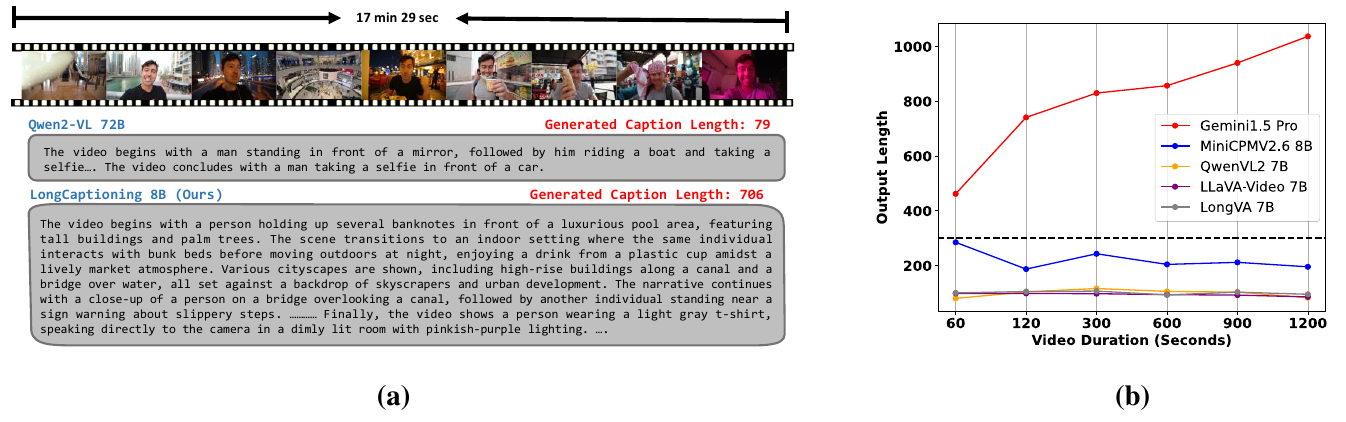}
	\caption{The output length of LMMs varies with different duration. 
	      The maximum output length of open-source LMMs is around 300 words, which is significantly shorter than that of proprietary models.}
	\label{fig:exp1}
  \end{figure*}

To evaluate the performance of current LMMs on long video captioning, we design a set of experiments specifically tailored to this task. 
In particular, we collect 5 movie videos, each approximately 20 minutes long, and segment each video into 6 clips of 1, 2, 5, 10, 15, and 20 minutes. 
For videos of different durations, we sample one frame every 4 seconds as input to the model. 
This ensures that as the video duration increases, the amount of information it contains also becomes richer. 
We then use the same prompts (detailed prompts are provided in the Appendix) to have different models perform the video captioning task. 
\textbf{We find that although the amount of visual information in the videos increases with length, open-source LMMs consistently struggle to generate outputs exceeding approximately 300 words, even as video duration and complexity grow.} 
This limitation results in a loss of critical visual details, as the models fail to capture or adequately describe the full scope of content. 
For example, as shown in Figure~\ref{fig:exp1} (a), Qwen-VL 72B generated only 79 words for a 17-minute video, significantly omitting visual details. Figure~\ref{fig:exp1} (b) shows that other open-source models exhibit a similar pattern, with outputs consistently below 300 words even as video length increases. 
In contrast, the proprietary model (e.g., Gemini 1.5 Pro~\cite{team2024gemini}) performs exceptionally well, generating over 1,000 words for a 20-minute video, effectively preserving rich, fine-grained information. However, given the lack of open-source solutions for such proprietary models and the deployment challenges they pose on edge devices, we aim to explore a low-cost method to enhance the long-form video caption generation capabilities of open-source LMMs.

LMMs typically consist of a visual encoder, a modality projector, and a language decoder, where the language decoder is usually a pretrained large language model (LLM). 
As a core component, the development of LLMs often drives advancements in LMMs. LongWriter~\cite{bai2024longwriter} is the first to explore the long text generation problem in LLMs. 
By constructing a supervised fine-tuning dataset for long texts, it successfully extended the model output from 2,000 to 10,000 words. 
Inspired by this work, we design a set of controlled experiments to investigate long caption generation in the context of long video captioning. 
We find that despite the increasing volume of visual information with longer videos, open-source LMMs consistently fail to generate captions exceeding 300 words. 
\textbf{The root cause of this limitation lies in the scarcity of long-caption examples during training, which hinders the model’s ability to generate outputs beyond a certain length.} 
This issue becomes particularly pronounced as the video duration increases, as the model’s context window is insufficient to capture all relevant details. 
To overcome this limitation, one potential solution is to construct datasets of video-long caption pairs. However, manually annotating such datasets is time-consuming and expensive, which presents a significant barrier to improving long-caption generation in LMMs.

To address this, we propose the \textbf{LongCaption-Agent}, an automated framework for synthesizing long-caption data. 
This framework utilizes off-the-shelf LMMs and LLMs to synthesize captions in three stages: frame-level, clip-level, and video-level. 
By employing multi-level information extraction and summarization, the framework produces comprehensive long-captions. 
Building upon the LongCaption-Agent, we develop the \textbf{LongCaption-10K} dataset, which includes 10,000 long-caption examples. 
By incorporating the LongCaption-10K dataset into training, we extend the output length of LMM to over 1,000 words, successfully unlocking its long video caption generation capabilities. 
Additionally, to further enable the model to handle longer sampled frame inputs during inference, we introduce a visual context window extension method~\cite{wei2024visual} in the training phase. 
This method effectively mitigates the issue of reduced output length when transitioning from short-sequence training to long-sequence inference.

Traditional $n$-gram-based metrics ($e.g.$, CIDEr~\cite{VedantamZP15}) commonly used for image and video captioning fall short for long captions due to the flexibility of language. 
To rigorously evaluate the long-caption generation capabilities of LMM, we develop \textbf{LongCaption-Bench}, a comprehensive benchmark for assessing the quality of long-captions generated by LMMs. 
It includes 281 test videos with an average duration of 1060.4 seconds. 
Each video has a global description generated by Gemini 1.5 Pro~\cite{team2024gemini}, which is then manually reviewed and modified. 
The average caption length is 1161.3 words. 

Evaluations on LongCaption-Bench indicate that our model achieved optimal performance, even surpassing larger proprietary models. 
Additionally, to reduce training costs, we introduce a visual context window expansion technique~\cite{wei2024visual} during the training phase to further increase the effective context length during inference.
In summary, our paper makes the following key contributions:
\begin{itemize}
\item We are the first to explore the main factor limiting the output length of LMMs: the scarcity of long-caption examples in training data. 
\item To unlock the long-caption generation capability of LMMs at a lower cost, we propose the LongCaption-Agent, a long-caption synthesis framework. 
Based on this framework, we construct the LongCaption-10K. By incorporating the LongCaption-10K dataset into training, we extend the output length of LMM to over 1,000 words.
\item We also develop LongCaption-Bench, designed to comprehensively evaluate the quality of long-captions generated by LMMs. 
\end{itemize}

The rest of the paper is organized as follows. 
In Section~\ref{sec:rela_work}, we provide an overview of related work. 
Section~\ref{sec:longagent} presents a detailed description of the data synthesis process for the LongCaption-Agent and the statistical characteristics of LongCaption-10K dataset. 
In Section~\ref{sec:longcaption-bench}, we introduce LongCaption-Bench, the evaluation benchmark for long video captioning. 
Section~\ref{sec:longcaptioning} elaborates on the experimental setup and training details of LongCaptioning-8B, followed by an analysis of the experimental results. 
Finally, Section~\ref{sec:conclusion} provides a summary of the paper.
\begin{figure*}[t]
	\centering
	\includegraphics[width=0.94\linewidth]{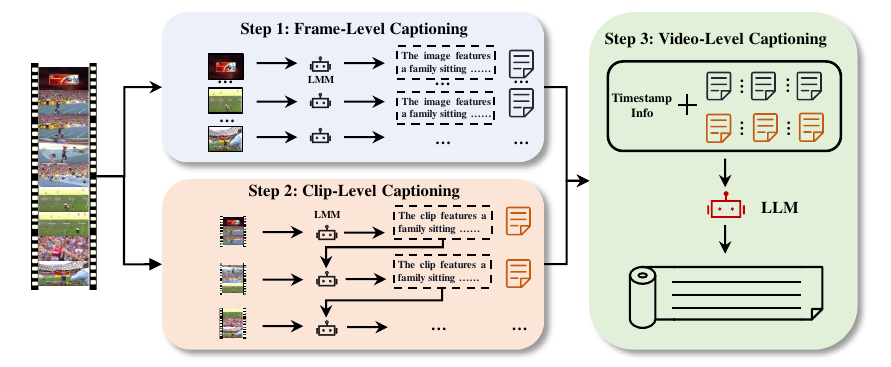}
	\caption{The LongCaption-Agent framework. 
	The framework uses an off-the-shelf LMM ($i.e.$, MiniCPMV2.6-8B~\cite{yao2024minicpm}) to first generate frame-level descriptions for each sampled frame, and then iteratively produce clip-level descriptions for each clip. 
	Finally, an off-the-shelf LLM ($i.e.$, GLM4-Long~\cite{glm2024chatglm}) is used to integrate the frame-level and clip-level descriptions into a complete video long-caption, incorporating additional context such as timestamps.}
	\label{fig:longagent}
  \end{figure*}

\section{Related Work}
\label{sec:rela_work}
\subsection{Large Multimodal Model}
Recent advancements in large language models (LLMs)~\cite{touvron2023llama,jiang2023mistral,yang2024qwen2,Claude3,gpt4o,achiam2023gpt} have demonstrated impressive language understanding and generation capabilities. 
This success has sparked interest in large multimodal models (LMMs)~\cite{li2024llava,Dai2023InstructBLIPTG,Ye2023mPLUGOwlME,yao2024minicpm}, which typically consist of visual encoders, modality projectors, and pretrained language model decoders. 
LMMs initially made breakthroughs in image understanding tasks. 
With the construction of high-quality video-text datasets, more researchers are applying LMMs to video understanding tasks~\cite{zhang2024long,xue2024longvila,liu2024kangaroo,li2024llava,li2023llamavid,jin2023chat-univi,song2024moviechat,xu2024pllava,lin2023video}. 
For example, models like VideoChatGPT~\cite{0001RKK24}, VideoChat~\cite{li2023videochat}, and Video-LLaMA~\cite{cheng2024videollama} have enhanced the video understanding capabilities of LMMs through high-quality data and fine-tuning techniques.  
These models have shown excellent performance in short video understanding. 
Recently, some efforts~\cite{zhang2024long,wei2024visual,xue2024longvila,liu2024kangaroo} have been made to input long videos into LMMs, achieving some progress. 
For instance, MovieChat~\cite{song2024moviechat} introduced a memory mechanism to compress long video tokens into a fixed size. 
Additionally, LongVA~\cite{zhang2024long} extended the context window by continuously training LLMs on long texts. 
Although they perform well in long video-QA tasks, they face challenges in generating video captions that require global descriptions. 
For a 20-minute video, they struggle to output even 300 words of description. 
This limits the application of the model in video understanding. 

\subsection{Video Captioning}
Early video captioning methods used template-based approaches~\cite{kojima2002natural,guadarrama2013youtube2text,krishnamoorthy2013generating}, which lacked flexibility. 
With the development of deep learning, expert models based on CNN-RNN and Transformer architectures replaced previous methods~\cite{gao2020fused,hu2019hierarchical}. However, these approaches typically handle only short videos of a few seconds, and the generated captions are also brief. 
The VideoReCap~\cite{islam2024video} model was the first to attempt generating descriptions for long videos recursively. 
However, its descriptions often do not exceed 100 words for a 60-minute video, inevitably missing much of the video content. 
Dense video captioning~\cite{YangNSMPLSS23,KimKMC024,WangZLZC021} typically identifies different event timestamps within a video and generates corresponding captions for each event. 
However, these methods still focus on short video-short caption scenarios, with annotated captions generally limited to no more than 30 words. 
Recently, some efforts~\cite{chen2024panda,chen2024sharegpt4video} have attempted to combine LMMs and LLMs to construct large-scale video captioning datasets. 
For instance, Panda70M~\cite{chen2024panda} constructed semantically consistent videos by splitting and merging based on semantic understanding, and then used a pre-trained model to generate captions for each video. 
ShareGPT4Video~\cite{chen2024sharegpt4video} proposed a differential video captioning strategy, leveraging GPT-4V~\cite{achiam2023gpt} to synthesize video captions by identifying differences between adjacent frames. 
However, these methods typically focus on short video-caption examples.

\section{LongCaption-Agent: A Long-Caption Synthesis Framework}
\label{sec:longagent}
In this section, we introduce LongCaption-Agent, a framework designed for generating long-captions for videos. 
Based on LongCaption-Agent, we develop LongCaption-10k long-caption dataset, which includes 10,000 long-caption examples. 
Next, we provide a detailed explanation of the data synthesis process and the statistical information of the dataset.
\begin{figure*}[t]
	\centering
	\begin{minipage}{0.5\textwidth}
	  \caption{Existing Video-Text Datasets}
	  \centering
	  \label{table:it_dataset_info}
	  \scalebox{0.8}{
	  \begin{tabular}{lcc}
	  \toprule
	  Dataset Name & Video Length (sec.) & Text Length  \\
	  \midrule
	  MSVD~\cite{ChenD11}  & 9.7 & 4.7 words \\
	  YouCook2~\cite{ZhouXC18}  & 19.6 & 8.8 words \\
	  MSR-VTT~\cite{XuMYR16}  & 15.0 & 9.3 words \\
	  ActivityNet~\cite{HeilbronEGN15} & 36.0 & 13.5 words \\
	  LLaVA-Hound-255K~\cite{zhang2024direct} & 52.4 & 37.6 words \\
	  VideoChatGPT-100K~\cite{0001RKK24} & 123.4 & 68.0 words \\
	  Panda-70M~\cite{chen2024panda} & 8.5 & 13.2 words \\
	  \hline
	  \textbf{LongCaption-10K (Ours)} & 92.8 & 1198.2 words \\
	  \bottomrule
	  \end{tabular}
	  }
	\end{minipage}
	\hfill
	\begin{minipage}{0.48\textwidth}
		\centering
		\includegraphics[width=0.85\linewidth]{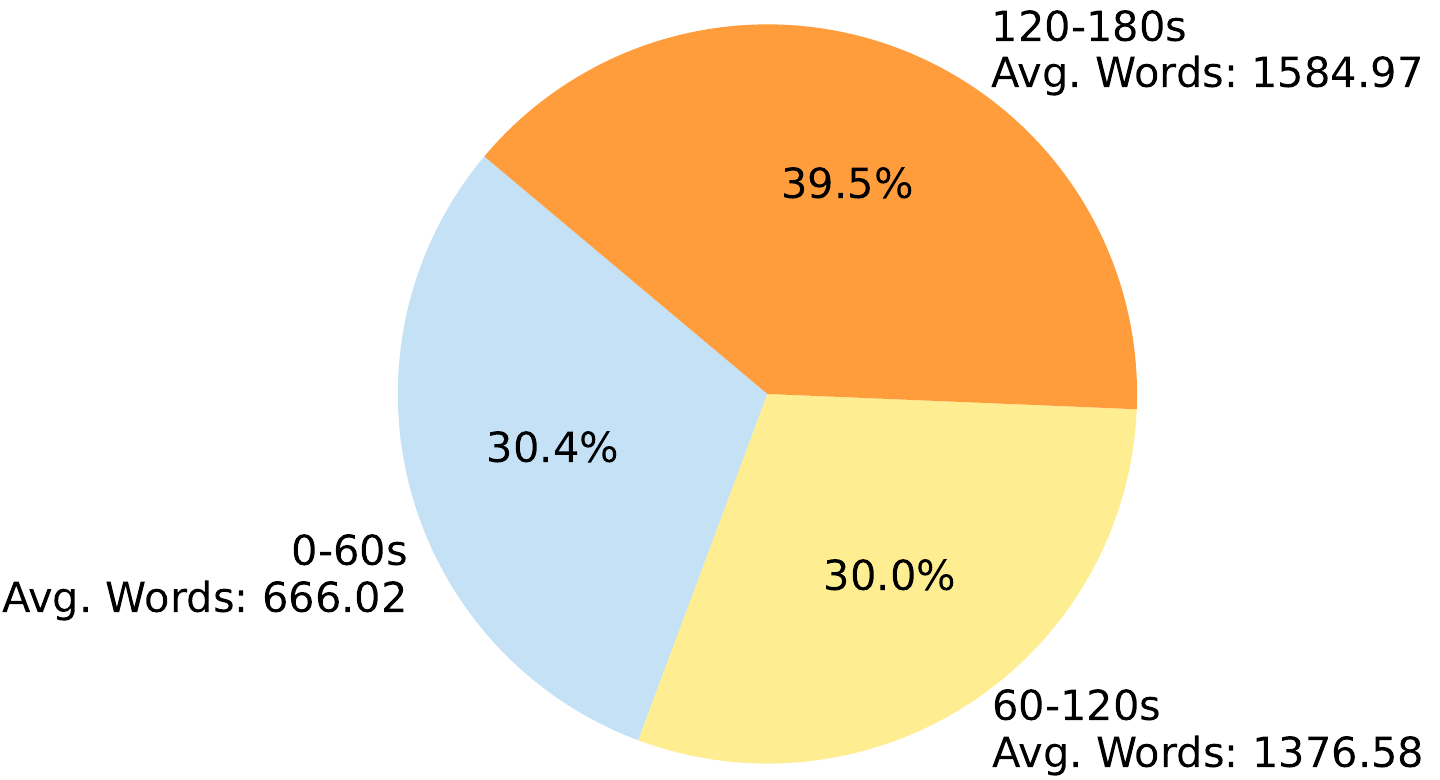}
		\caption{Key statistics of LongCaption-10k.}
		\label{fig:longcaption-10k}
	\end{minipage}
  \end{figure*}

\subsection{Framework}
In Section~\ref{sec:intro}, we conduct a detailed analysis of the challenges faced by previous large multimodal models (LMMs) in generating long-captions. Our investigation reveals that the primary factor contributing to these challenges is the scarcity of long-caption examples during the training phase. 
We conduct a statistical analysis of the commonly used video-text datasets. 
Table~\ref{table:it_dataset_info} shows the average video duration and average text length of these datasets, with all of their average text lengths being less than 100. 
Among them, Pand70M~\cite{chen2024panda} uses off-the-shelf open-source multimodal models to synthesize video caption data. 
Specifically, it employs multiple multimodal teacher models to generate captions for different modality combinations, and then uses a trained retrieval model to select the caption that best matches the video. 
However, due to the limitation of the model's output length, the captions synthesized by Pand70M are relatively short. 
In addition, using proprietary large multimodal models (e.g., Gemini 1.5 Pro~\cite{team2024gemini}) to synthesize long-captions, as well as manually annotating long-captions, are both extremely costly.

To obtain long-caption samples at a lower cost, we propose the \textbf{LongCaption-Agent}, a framework that synthesizes long-caption data by aggregating multi-level descriptions. 
Specifically, the framework leverages off-the-shelf open-source LMMs and LLMs to divide the long-caption synthesis process into three steps: 
1) Frame-level captioning: The LMM is used to extract static fine-grained information from each sampled frame. 
2) Clip-level captioning: The video is divided into multiple clips, and the LMM is employed to extract temporal fine-grained information from each short clip. 
3) Video-level captioning: Finally, with the powerful language understanding capabilities of the LLM, the frame-level and clip-level captions are aggregated to synthesize a complete video-level long-caption. 
Next, we provide a detailed introduction for each step.

First, we sample the video into a sequence of frames at 1 fps, and then generate captions at different levels: 

\noindent \textbf{Step 1 - Frame-Level Captioning}: 
Considering that current open-source LMMs struggle to generate long-captions for videos, we instead sample frames as model inputs to extract fine-grained static information from each frame. 
As shown in Figure~\ref{fig:longagent}, in this work, we use MiniCPMV2.6-8B~\cite{yao2024minicpm} as the frame-level caption generation model. 
Some studies have attempted to use LLMs to summarize frame-level captions into video captions. 
However, by neglecting the temporal information of the original video, this approach often fails to accurately capture the dynamic changes in events and the continuity of actions, resulting in captions that lack coherence and completeness. 
Temporal information is crucial for understanding causal relationships, action sequences, and scene transitions in videos. 
Relying solely on frame-level summaries tends to produce fragmented captions that struggle to reflect the overall semantics of the video. 
Therefore, in this work, we use frame-level captions only as a supplementary source of fine-grained static information for clip-level captions.

\noindent \textbf{Step 2 - Clip-Level Captioning}: 
Due to the issue of the lack of coherence and consistency when directly integrating frame-level captions, we introduce clip-level captioning that incorporates temporal information. 
Specifically, we first divide the original video into multiple clips. 
To avoid excessive content disparity between the captions of different clips and ensure smooth transitions between frame-level captions, we designed a clip segmentation method based on a sliding window. 
Specifically, we extract clips using a sliding window approach, where the window size is set to 10 seconds and the stride is set to 5 seconds, thereby constructing overlapping clips that share content from adjacent time segments. 
This overlapping design not only captures the continuity of the video but also effectively mitigates the issue of context fragmentation caused by direct segmentation. 
In this way, we generate more coherent clip-level captions while preserving the temporal information of the video.

Moreover, to further enhance the coherence and consistency of caption generation, we design an iterative caption generation strategy. Specifically, when generating the caption for each clip, we not only rely on the visual information of the current clip but also input the caption generated for the previous clip as contextual information into LMM. This allows the model to leverage contextual information from both the preceding and following clips, generating captions that better align with the overall semantics of the video and avoiding content jumps or breaks between different clip captions. 
This iterative generation process is represented by the following formula:
\begin{equation}
  \operatorname{caption}_{t} = \operatorname{LMM}(\operatorname{clip}_{t}, \operatorname{caption}_{t-1})
\end{equation}
where $\operatorname{clip}_{t}$ represents the clip at time step $t$ that requires captioning, and $\operatorname{caption}_{t-1}$ denotes the caption generated for the clip at time step $t-1$.

Considering that a simple summary of clip-level captions may overlook some fine-grained static information from sampled frames, we propose using LLM to integrate the frame-level captions, which contain fine-grained static information, with the clip-level captions, which incorporate temporal information.
\begin{figure*}[t]
	\centering
	\includegraphics[width=0.90\linewidth]{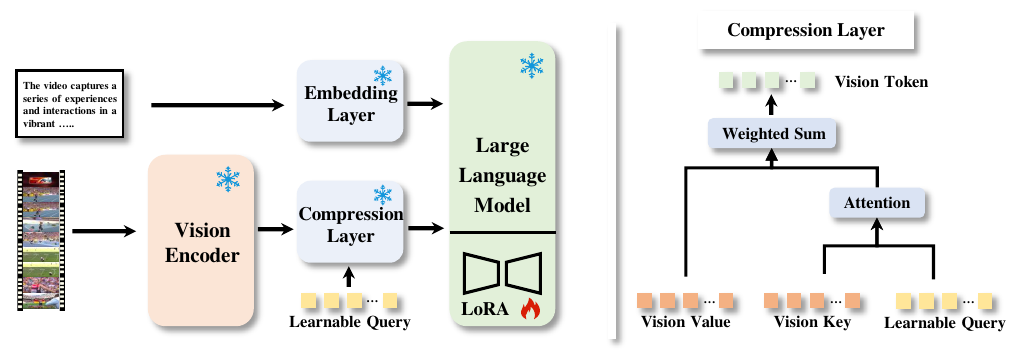}
	\caption{The framework of the Large Multimodal Model (LMM). 
  The model consists of a vision encoder that processes video frames and generates vision tokens. 
  These tokens are then passed through a compression layer that reduces the number of vision tokens. 
  The compression layer uses weighted sum and attention mechanisms, incorporating vision tokens, and learnable queries tokens, where the number of learnable queries is much smaller than the number of value and key. 
  The processed information is fed into a Large Language Model (LLM), which leverages LoRA (Low-Rank Adaptation) to train language model.}
	\label{fig:lmm}
\end{figure*}

\noindent \textbf{Step 3 - Video-Level Captioning}: 
With the emergence of large language models such as ChatGPT~\cite{ChatGPT}, LLMs have demonstrated remarkable capabilities in language understanding and generation. Therefore, we have designed an LLM-based caption summarization pipeline that, through carefully crafted prompts, can efficiently and accurately integrate frame-level descriptions into clip-level captions and summarize all clip-level captions into a complete video-level long-caption.

Specifically, our pipeline is divided into the following steps: 
1) Timestamp Annotation: First, we annotate each clip-level caption and each frame-level caption with a timestamp, allowing the LLM to align the frame-level descriptions with the clip-level captions in terms of time. 
2) Frame-Level Caption Grouping: Next, based on the timestamps of clip-level captions, we group the continuous frame-level captions into different sets, with each set corresponding to the time range of a clip. 
This ensures that the frame-level captions can be summarized within a local time range without losing details. 
3) Interleaved Input Combination: Finally, we interleave all the frame-level caption sets with the clip-level captions and, combined with carefully designed prompts, feed them as input to the LLM. 
The prompts are designed to guide the LLM in effectively merging the fine-grained information from the frame-level captions with the temporal information from the clip-level captions, thereby generating a coherent and information-rich long-caption.

Through this method, we not only retain the fine-grained static information from the frame-level captions but also ensure the temporal coherence of the video-level captions, ultimately generating a complete and rich  long-caption.
We provide the detailed prompts used at each step in Appendix.

\subsection{LongCaption-10K Dataset}
Based on the LongCaption-Agent, we constructed LongCaption-10k, a dataset containing 10,000 synthesized long-caption samples. 
Specifically, we selected 10,000 videos from open-source video datasets~\cite{zhang2024video}, with video durations ranging from 30 to 180 seconds. 
To ensure the model's generalization across different instructions, we collected a variety of prompts to serve as queries for LongCaption-10k. 

Table~\ref{table:it_dataset_info} reports the comparison results of video duration and annotation text length between LongCaption-10K and other open-source datasets. 
Among them, VideoChatGPT-100K contains longer videos, but its average text length is only 68 words. 
The average text length in the Panda70M dataset, which is based on LMM methods, is only 13.2. 
In contrast, LongCaption-10K extends the average annotation text length to 1,198 words, with an average video duration of 92.8 seconds. 
Figure~\ref{fig:longcaption-10k} further illustrates the distribution of samples with different video durations in the dataset. 
The LongCaption-10K dataset contains video data of varying durations and is relatively evenly distributed across the three intervals: [0, 60), [60, 120), and [120, 180). 
Additionally, across different duration ranges, this dataset consistently maintains a relatively longer caption length.

\begin{table}[!tbp]
  \centering
  \caption{Key statistics of LongCaption-Bench.}
  \label{tab:longcaption-bench}
  \resizebox{0.90\linewidth}{!}{
  \begin{tabular}{l|c|l|c}
  \hline
  \multicolumn{4}{c}{\textbf{\# Data in each subset}} \\ \hline
  Video duration & number & Caption length & number \\ \hline
  {[}300s, 600s) & 9 & {[}0, 500) & 38 \\ 
  {[}600s, 900s) & 58 & {[}500, 1000) & 101 \\ 
  {[}900s, 1200s) & 129 & {[}1000, 1500) & 66 \\ 
  {[}1200s, 1800s{]} & 85 & {[}1500, 3000) & 76 \\ \hline
  \multicolumn{2}{l|}{Average video duration} & \multicolumn{2}{c}{1060.4} \\ \hline
  \multicolumn{2}{l|}{Average caption length} & \multicolumn{2}{c}{1161.3} \\ \hline
  \end{tabular}}
\end{table}
\begin{table*}[t]
	\centering
	\caption{Evaluatio results of the video-caption relevance score on the LongCaption-Bench across different duration intervals, where ``Overall'' refers to the average score for the entire dataset, ``[300s, 600s)'' refers to the score for videos with duration between 300 seconds and 600 seconds, ``[600s, 900s)'' refers to the score for videos with duration between 600 seconds and 900 seconds, and ``[900s, 1200s)'' refers to the score for videos with duration between 900 seconds and 1200 seconds, and ``[1200s, 1800s)'' refers to the score for videos with duration between 1200 seconds and 1800 seconds. The score is calculated by averaging the scores of each video in the corresponding duration interval.}
	\resizebox{0.70\linewidth}{!}{
	\begin{tabular}{l|c|c|c|c|c}
	\toprule
	 & \multicolumn{1}{c|}{\textbf{Overall}} & \multicolumn{1}{c|}{\textbf{[300s, 600s)}} & \multicolumn{1}{c|}{\textbf{[600s, 900s)}} & \multicolumn{1}{c|}{\textbf{[900s, 1200s)}} & \multicolumn{1}{c}{\textbf{[1200s, 1800s)}} \\
	\midrule
	\multicolumn{3}{l}{\emph{Proprietary models}} \\
	\textbf{GPT-4o}~\cite{gpt4o} & 2.50  & 2.78  & 2.41  & 2.57  & 2.50 \\
	\midrule
	\multicolumn{3}{l}{\emph{Open-source models}} \\
	\textbf{PLLaVA 7B}~\cite{xu2024pllava}  & 1.10  & 1.09  & 1.07  & 1.10 & 1.18 \\
	\textbf{PLLaVA 32B}~\cite{xu2024pllava}  & 1.18  & 1.21  & 1.10  & 1.20  & 1.17 \\
	\textbf{LongVA 7B}~\cite{zhang2024long}   & 1.15  & 1.11  & 1.07 & 1.17   & 1.18\\
	\textbf{LLaVA-Video 7B}~\cite{liu2024llavanext}  & 1.30 & 1.33  & 1.21  & 1.38  & 1.24 \\
	\textbf{MiniCPMV2.6-8B}~\cite{yao2024minicpm}   & 2.18  &2.11   &2.02   &2.24  &2.20 \\
	\textbf{Qwen2-VL 7B}~\cite{wang2024qwen2}  & 1.36  & 1.22   & 1.12 & 1.47  & 1.36 \\
	\textbf{Qwen2-VL 72B}~\cite{wang2024qwen2}   & 1.71  & 1.72  & 1.66  & 1.72  & 1.68 \\
	\midrule
	\multicolumn{3}{l}{\emph{Our trained models}} \\
	\textbf{LongCaptioning-8B}   & \textbf{2.59}   & \textbf{2.56}  & \textbf{2.53}  & \textbf{2.62}  & \textbf{2.58} \\
	\bottomrule
	\end{tabular}
	}
	\label{tb:relevance_quality}
  \end{table*}

\begin{table*}[t]
  \centering
  \caption{Evaluation results of the length score $S_l$ and the quality score $S_q$ on the LongCaption-Bench across different duration intervals. Here, $S_l$ represents the score for caption length, $S_q$ represents the quality score.}
  \resizebox{0.80\linewidth}{!}{
  \begin{tabular}{l|cc|cc|cc|cc|cc}
  \toprule
   & \multicolumn{2}{c|}{\textbf{Overall}} & \multicolumn{2}{c|}{\textbf{[300s, 600s)}} & \multicolumn{2}{c|}{\textbf{[600s, 900s)}} & \multicolumn{2}{c|}{\textbf{[900s, 1200s)}} & \multicolumn{2}{c}{\textbf{[1200s, 1800s)}} \\
   \cmidrule(lr){2-3} \cmidrule(lr){4-5} \cmidrule(lr){6-7} \cmidrule(lr){8-9} \cmidrule(lr){10-11}
    & $S_l$ & $S_q$    & $S_l$ & $S_q$ & $S_l$ & $S_q$  & $S_l$ & $S_q$  & $S_l$ & $S_q$ \\
  \midrule
  \multicolumn{3}{l}{\emph{Proprietary models}} \\
  \textbf{GPT-4o}~\cite{gpt4o} & 33.6 & 79.4 & 64.9 & 81.5  & 41.2 & 81.2  & 30.2 & 79.4  & 24.3  & 78.7  \\
  \midrule
  \multicolumn{3}{l}{\emph{Open-source models}} \\
  \textbf{PLLaVA 7B}~\cite{xu2024pllava} & 4.5 & 63.2  & 4.0 & 64.1  & 2.3 & 63.7  & 3.7 & 64.9  & 6.8  & 62.8 \\
  \textbf{PLLaVA 32B}~\cite{xu2024pllava} & 5.3 & 65.1 & 4.9 & 66.4  & 2.6 & 64.7  & 4.9 & 65.2 & 7.7  & 64.0 \\
  \textbf{LongVA 7B}~\cite{zhang2024long}  & 2.1 & 64.5  & 0.0 & 65.2  & 0.0 &65.5  & 1.9 & 64.6  & 4.0  & 63.6  \\
  \textbf{LLaVA-Video 7B}~\cite{liu2024llavanext} & 4.5 & 66.8  & 8.6 & 67.8  & 1.8 & 69.0  & 3.1 & 67.2  & 8.4  & 64.8 \\
  \textbf{MiniCPMV2.6-8B}~\cite{yao2024minicpm}  & 11.8 & 73.5  & 27.6 & 72.6  & 12.8 & 74.9   & 10.4 & 73.1  & 11.5  & 73.0 \\
  \textbf{Qwen2-VL 7B}~\cite{wang2024qwen2} & 7.7 & 57.8  & 0.0 & 58.9  & 6.6 & 56.4  & 6.4 & 58.5 & 11.2  & 57.6  \\
  \textbf{Qwen2-VL 72B}~\cite{wang2024qwen2}  & 9.4 & 70.2  & 11.8 & 69.1  & 7.7 & 68.6  & 7.0 & 71.4  & 10.1  & 70.2   \\
  \midrule
  \multicolumn{3}{l}{\emph{Our trained models}} \\
  \textbf{LongCaptioning-8B}  & \textbf{40.9} & \textbf{81.2}  & \textbf{88.2} & \textbf{80.7}   & \textbf{53.6} & \textbf{81.7}  & \textbf{38.8} & \textbf{81.6}   & \textbf{30.3}  & \textbf{80.4}  \\
  \bottomrule
  \end{tabular}
  }
  \label{tb:length_quality}
\end{table*}
\begin{figure*}[t]
  \centering
  \begin{minipage}{0.24\textwidth}
      \centering
      \includegraphics[width=\linewidth]{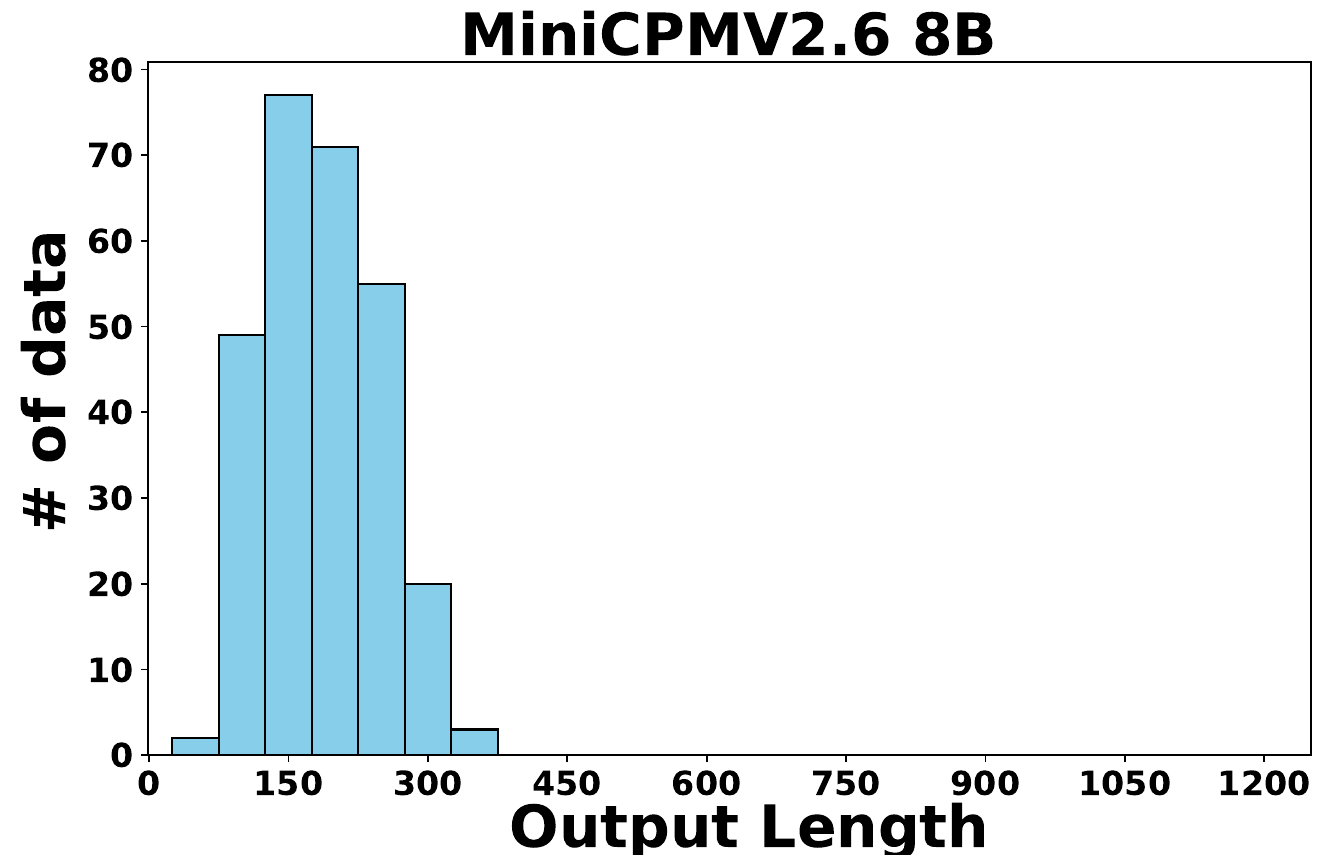}
      \label{fig:len_minicpm}
  \end{minipage}%
  \begin{minipage}{0.24\textwidth}
      \centering
      \includegraphics[width=\linewidth]{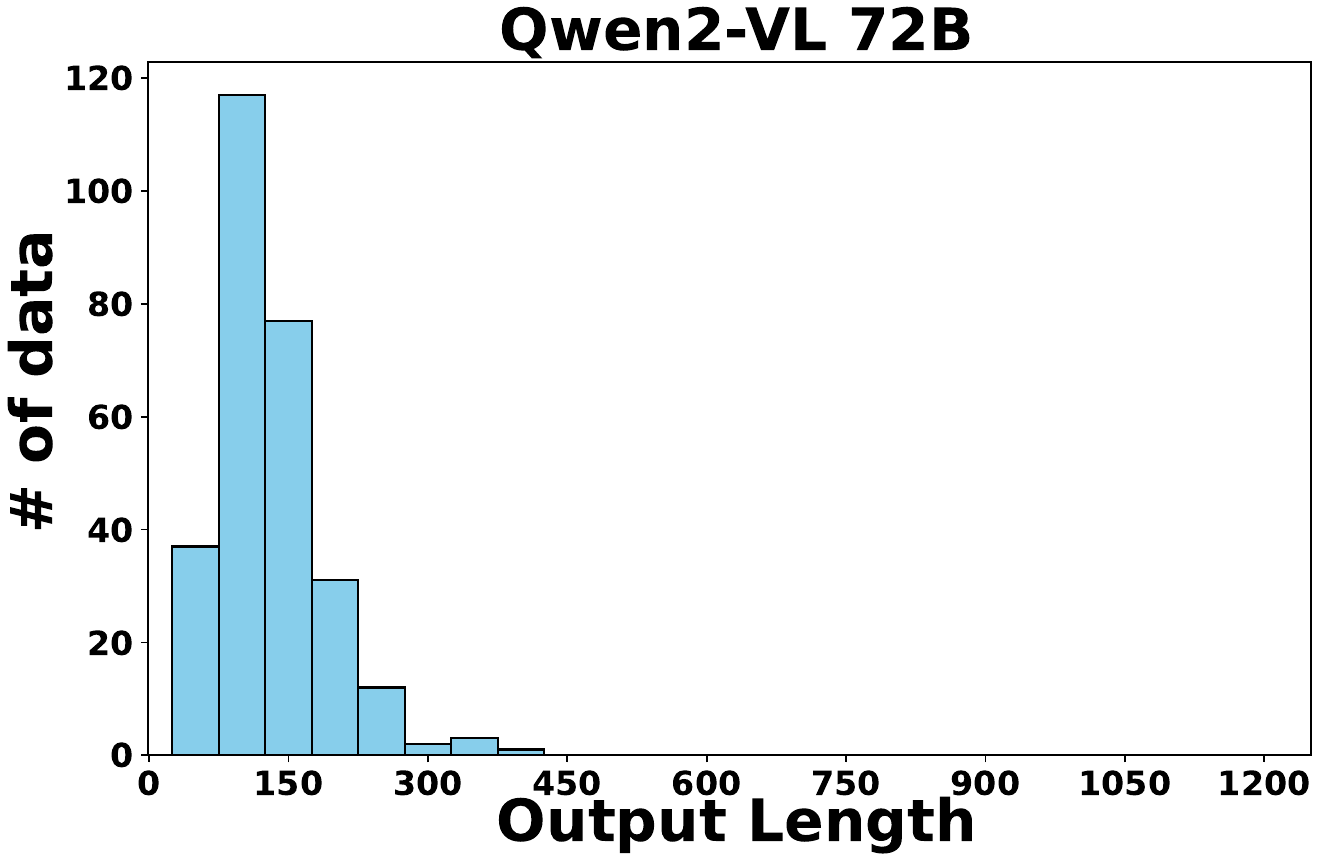}
      \label{fig:len_qwen72}
  \end{minipage}
  \begin{minipage}{0.24\textwidth}
    \centering
    \includegraphics[width=\linewidth]{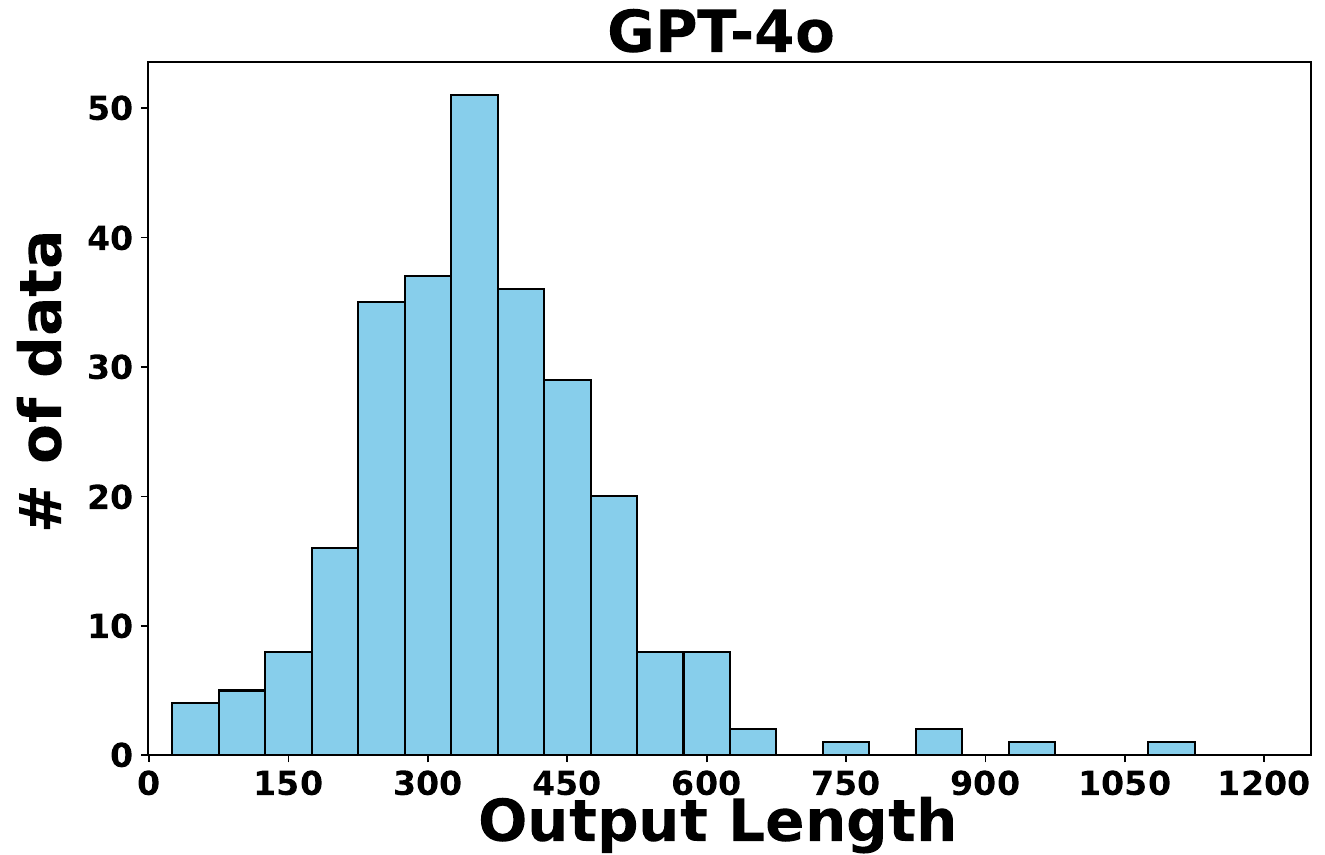}
    \label{fig:len_gpt}
\end{minipage}
\begin{minipage}{0.24\textwidth}
  \centering
  \includegraphics[width=\linewidth]{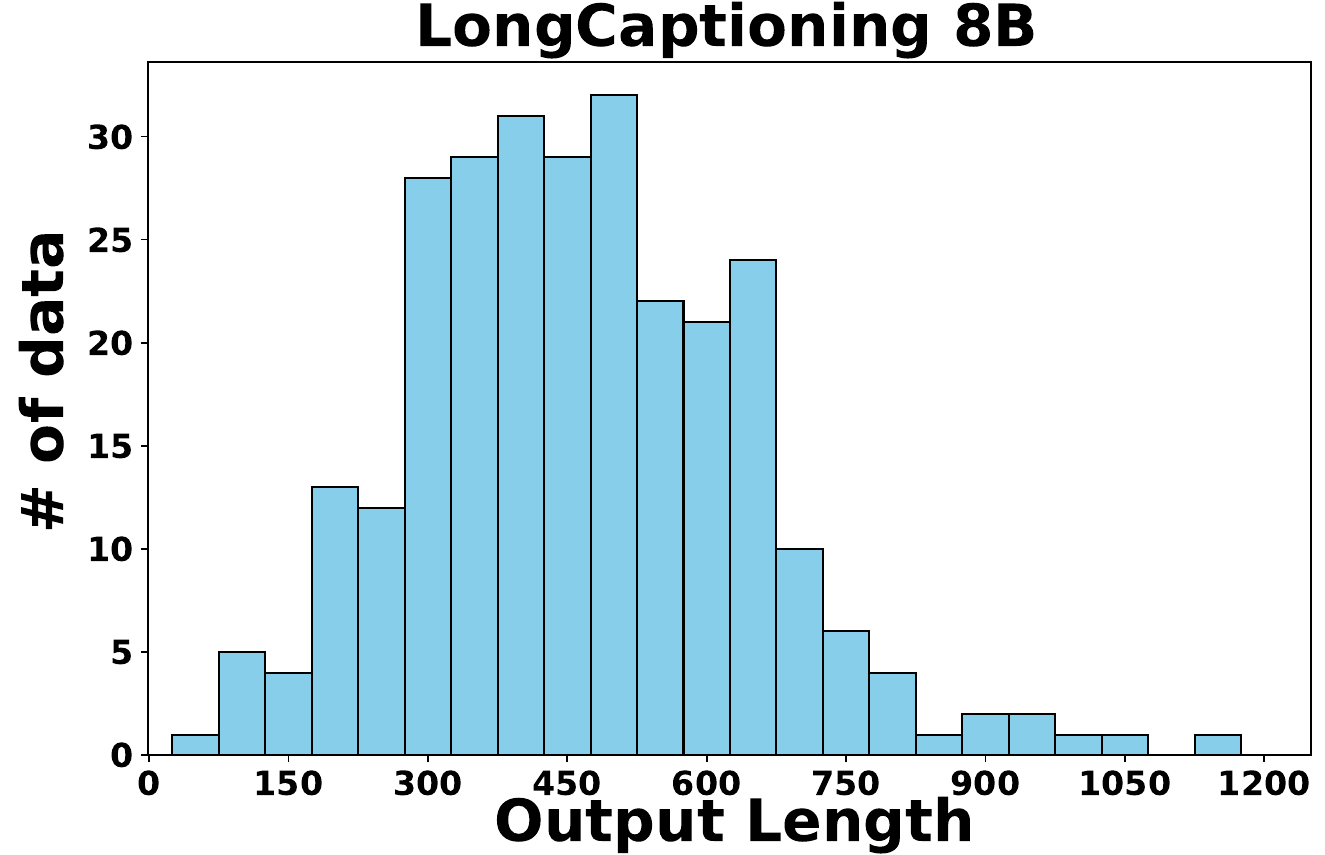}
  \label{fig:len_our}
\end{minipage}
\caption{Visualization of caption lengths generated by different models. The horizontal axis represents caption length, while the vertical axis represents the number of samples.}
\label{fig:length_compare}
\end{figure*}

\begin{table*}[t]
	\centering
	\caption{Ablation study results of the video-caption relevance score on the LongCaption-Bench. 
	`\emph{w/o Visual Context Window Ext.}' represents the LongCaptioning 8B model trained without the visual context window extension, 
	`\emph{w/o Clip-Level Captioning}' indicates that only frame-level captions were used during long caption synthesis, 
	`\emph{w/o Frame-Level Captioning}' refers to the use of only clip-level captions during long caption synthesis, 
	and `\emph{w/o LongCaption-10k data}' refers to the results without using the LongCaption-10K dataset for training.}
	\resizebox{0.85\linewidth}{!}{
	\begin{tabular}{l|c|c|c|c|c}
	\toprule
	& \multicolumn{1}{c|}{\textbf{Overall}} & \multicolumn{1}{c|}{\textbf{[300s, 600s)}} & \multicolumn{1}{c|}{\textbf{[600s, 900s)}} & \multicolumn{1}{c|}{\textbf{[900s, 1200s)}} & \multicolumn{1}{c}{\textbf{[1200s, 1800s)}} \\
	\midrule
	\textbf{LongCaptioning-8B}    & \textbf{2.59}  & \textbf{2.56} & \textbf{2.53}  & \textbf{2.62}  & \textbf{2.58} \\
	\quad \emph{w/o Visual Context Window Ext.}& 2.35 & 2.44 & 2.13 & 2.37 & 2.45\\
	\quad \emph{w/o Clip-Level Captioning}& 2.21 &2.25   &2.06   &2.28  &2.30 \\
	\quad \emph{w/o Frame-Level Captioning}& 2.20  &2.20  &2.06   &2.27  &2.28 \\
	\quad \emph{w/o LongCaption-10k data}  & 2.18 &2.11  &2.02  &2.24  &2.20 \\
	\bottomrule
	\end{tabular}
	}
	\label{tb:longcaptioning_ablation_relevance}
  \end{table*}

\begin{table*}[!ht]
  \centering
  \caption{Ablation study results of the length score $S_l$ and quality score $S_q$ on the LongCaption-Bench.}
  \resizebox{0.90\linewidth}{!}{
    \begin{tabular}{l|cc|cc|cc|cc|cc}
		\toprule
		 & \multicolumn{2}{c|}{\textbf{Overall}} & \multicolumn{2}{c|}{\textbf{[300s, 600s)}} & \multicolumn{2}{c|}{\textbf{[600s, 900s)}} & \multicolumn{2}{c|}{\textbf{[900s, 1200s)}} & \multicolumn{2}{c}{\textbf{[1200s, 1800s)}} \\
		 \cmidrule(lr){2-3} \cmidrule(lr){4-5} \cmidrule(lr){6-7} \cmidrule(lr){8-9} \cmidrule(lr){10-11}
		  & $S_l$ & $S_q$    & $S_l$ & $S_q$ & $S_l$ & $S_q$  & $S_l$ & $S_q$  & $S_l$ & $S_q$ \\
  \midrule
  \textbf{LongCaptioning-8B}  & \textbf{40.9} & \textbf{81.2}  & 88.2 & \textbf{80.7}  & \textbf{53.6} & \textbf{81.7}  & \textbf{38.8} & \textbf{81.6}  & \textbf{30.3}  & \textbf{80.4} \\
  \quad \emph{w/o Visual Context Window Ext.} & 32.3 & 79.1  & \textbf{90.3} & 76.3  & 37.5 & 78.7  & 30.1 & 79.7 & 25.8 & 78.6 \\
  \quad \emph{w/o Clip-Level Captioning} & 28.7 & 75.0 & 64.3 & 74.8 & 25.8 & 76.5   & 21.0 & 76.7   & 19.1  & 74.9  \\
  \quad \emph{w/o Frame-Level Captioning} & 25.7 & 74.2 & 58.8 & 74.6 & 21.5 & 76.1   & 18.9 & 76.0  & 16.0  & 74.3\\
  \quad \emph{w/o LongCaption-10k data} & 11.8 & 73.5  & 27.6 & 72.6 & 12.8 & 74.9   & 10.4 & 73.1 & 11.5  & 73.0\\
  \bottomrule
  \end{tabular}
  }
  \label{tb:longcaptioning_ablation_length}
\end{table*}
\begin{table}[t]
	\small
	\renewcommand{\arraystretch}{1.0}
	\begin{center}
	\caption{{Performance evaluation on VideoMME~\citep{fu2024video} benchmark, where ``Short'' refers to the duration of 0s to 120s, ``Medium'' refers to the duration of 240s to 900s, ``Long'' refers to the duration of 1800s to 3600s, and ``Overall'' refers to the average of all duration ranges.}}
	\resizebox{0.98\linewidth}{!}{
	\begin{tabular}{l|cccc}
	\toprule   \textbf{Methods}     & \textbf{Short}  & \textbf{Medium}  & \textbf{Long}  & \textbf{Overall} \\ 
	\midrule
		
    ST-LLM-7B~\citep{liu2024st}      & 45.7  & 36.8  & 31.3  & 37.9   \\ 
		VideoLLaVA-7B~\citep{lin2023video}     & 45.3  & 38.0  & 36.2  & 39.9 \\ 
		VideoChat2-Mistral-7B~\citep{li2024mvbench}   & 48.3  & 37.0  & 33.2  & 39.5    \\ 
    Chat-UniVi-V1.5-7B~\citep{jin2024chat}     & 45.7  & 40.3  & 35.8  & 40.6   \\ 
    Qwen-VL-Chat-7B~\citep{bai2023qwen}  & 46.9  & 38.7  & 37.8  & 41.1 \\ 
		VideoLLaMA2-7B~\citep{cheng2024videollama}    & 56.0  & 45.4  & 42.1  & 47.9   \\ 
		LLaVA-NeXT-Qwen2-7B~\citep{liu2024llavanext}   & 58.0  & 47.0  & 43.4  & 49.5   \\ 
    LongVA-7B~\citep{zhang2024long}  & 61.1 & 50.4  & 46.2  & 52.6 \\ 
		LLaVA-OneVision-7B~\citep{li2024llava}   & 69.3  & 55.1  & 49.7  & 58.2   \\ 
    MiniCPMV2.6-8B~\citep{yao2024minicpm}   & 71.3	& 59.4	& 51.8	& 60.9  \\
		\midrule
	 \textbf{LongCaptioning-8B} & \textbf{72.2}	& \textbf{59.9}	& \textbf{53.4}	& \textbf{62.4}  \\ 
		\bottomrule
	\end{tabular}
	}
	\label{tab:video-mme}
	\end{center}
\end{table}

\section{LongCaption-Bench: A Benchmark for Long-Caption Generation}
\label{sec:longcaption-bench}
Table~\ref{table:it_dataset_info} presents commonly used video captioning benchmarks, which are focused on short video-short caption scenarios. 
Although our synthesized dataset, LongCaption-10K,  provides long caption annotations, traditional video captioning benchmarks are typically annotated manually, and the reliability and quality of human-annotated samples generally surpass those of synthesized samples.

To reliably evaluate the long-caption generation capabilities of LMMs, we constructed a LongCaption-Bench. 
Specifically, to prevent test videos from being included in the training data, we selected videos from the test sets of open-source long video benchmarks, with video durations ranging from 5 to 30 minutes. 
To reduce the difficulty of manual annotation, we first used the latest proprietary LMM, Gemini 1.5 Pro~\cite{team2024gemini} (which supports long-caption generation), to generate an initial caption for all videos. 
We then manually reviewed each video and its corresponding initial caption, removing any abnormal captions and their associated samples. 
Abnormal captions included those with infinite repetition, incomplete captions (abrupt termination), and sensitive content. 
In the end, we retained 281 valid video-caption pairs. Subsequently, we manually reviewed the initial captions against the video content and corrected or supplemented any errors or omissions in the captions. 
In the appendix, we present the tool interface used for modifying and supplementing the captions. 
Table~\ref{tab:longcaption-bench} reports the statistics of LongCaption-Bench. The test videos are primarily concentrated in the 900 to 1,200-second range, with caption lengths mostly between 500 and 1,500 words. 
The average video duration is 1060.4s, and the average caption length is 1161.3 words, which is significantly greater than previous video captioning benchmarks. 

Another challenge in evaluating long-captions lies in the assessment metrics. 
Traditional $n$-gram-based metrics ($e.g.$, CIDEr~\cite{VedantamZP15}) commonly used for image and video captioning fall short for long captions due to the flexibility of language. 
In this work, we evaluate the long-caption generation capabilities of LMMs from three perspectives: length, quality, and video-caption relevance. 

\noindent \textbf{Length Score $S_l$}: 
Following~\cite{bai2024longwriter}, we use the length of human-annotated captions as a reference and calculate the model's output length score $S_l$ based on a piecewise linear function: 
\begin{equation}
  \label{eq:sl}
  S_l= \begin{cases}100 \cdot \max \left(0,1-\left(l^{\prime} / l-1\right) / 3\right) & \text { if } l^{\prime}>l \\ 
    100 \cdot \max \left(0,1-\left(l / l^{\prime}-1\right) / 2\right) & \text { if } l^{\prime} \leq l\end{cases}
\end{equation}
where $l$ is the human-annotated length and $l'$ is the model's output length. 
The piecewise linear function ensures that, compared to the ground truth, both excessively long or excessively short model outputs will result in lower scores.

\noindent \textbf{Quality Score $S_q$}: 
Considering that the comparative methods include GPT-4o, in order to ensure a fair comparison, we use GPT-4o mini~\cite{GPT-4o-mini} as the judge. 
Following~\cite{bai2024longwriter}, we score the output across six dimensions: relevance, accuracy, coherence, clarity, breadth and depth, and readability. 
We take the average score across these six dimensions to obtain the overall quality score. 


\noindent \textbf{Video-Caption Relevance Score $S_r$}: 
Following Video-ChatGPT~\cite{0001RKK24}, we score the correlation between human annotations and model-generated descriptions on a scale of 1 to 5. 
To ensure a fair comparison, we use the original prompts provided by Video-ChatGPT for evaluation. 

\section{LongCaptioning: A Long Video Caption Generation Model}
\label{sec:longcaptioning}
In this section, we discuss the training details of the long-caption generation model trained on the constructed LongCaption-10K dataset, as well as the corresponding experimental results.
\subsection{Model Training}
We conduct training based on the latest open-source models, namely MiniCPMV2.6-8B~\cite{yao2024minicpm}. 
Figure~\ref{fig:lmm} shows the structure of our model. 
This model has only undergone pretraining and supervised fine-tuning (SFT) on image-text datasets. 
This allows us to eliminate the interference of other video datasets when analyzing the experimental results. 
Moreover, to further extend the effective context window during inference and allow the input of longer frame sequences, we introduced a visual context window extension technique~\cite{wei2024visual} during training. 
This technique scales the rotational frequency of the position encoding embedding for visual tokens, thereby extending the model's context window. 
The model was trained on a 2 A800-80G GPUs using DeepSpeed + ZeRO3 + CPU offloading~\cite{AminabadiRALLZRSZRH22}. 
We used a batch size of 1, a learning rate of 5e-5, and trained for 1 epoch.
\begin{figure*}[!t]
	\centering
	\includegraphics[width=0.99\linewidth]{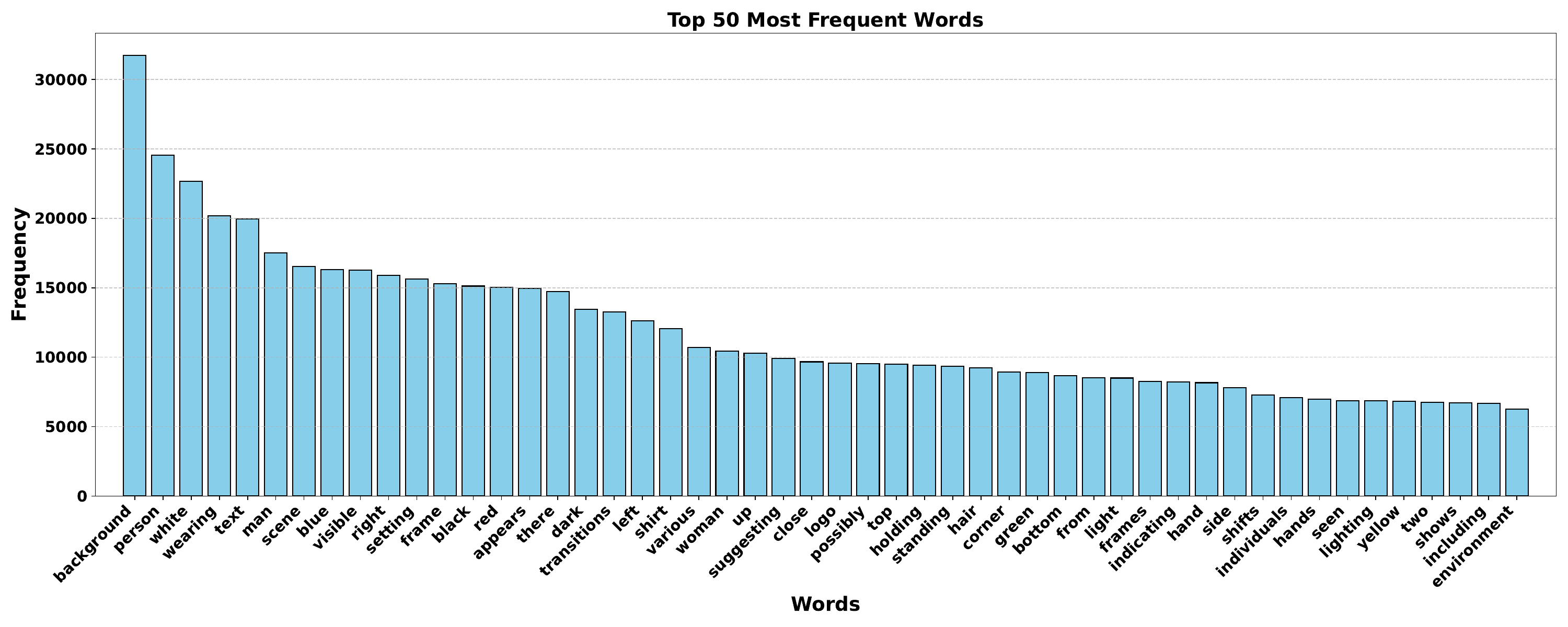}
	\caption{Statistical results of the top 50 words in LongCaption-10K.}
	\label{fig:top_100_words}
\end{figure*}
\begin{figure*}[!t]
	\centering
	\includegraphics[width=0.99\linewidth]{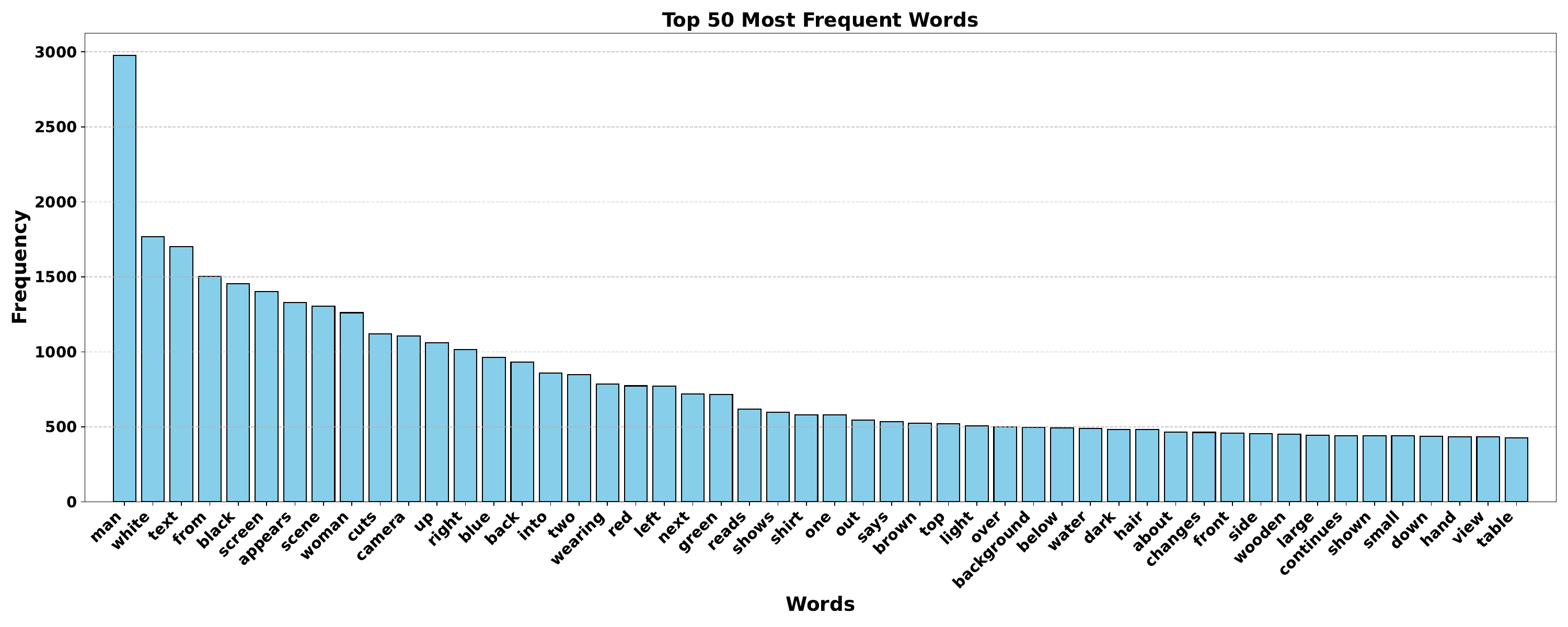}
	\caption{Statistical results of the top 50 words in LongCaption-Bench.}
	\label{fig:bench_top_100_words}
\end{figure*}

\subsection{Experiments}
We compare the LongCaptioning model with state-of-the-art open-source models and proprietary models on the LongCaption-Bench. 
For videos of different duration, we sample one frame every 6 seconds as input to the model. 
This ensures that as the video duration increases, the amount of information contained also becomes richer. 
It is worth noting that since the LongCaption-Bench was first generated by Gemini1.5 Pro and then refined by human editors, we do not use Gemini 1.5 Pro as a comparison model. 
For the open-source models, we uniformly set the output temperature to 0.2 and the maximum token generation parameter to 2048. 

\subsubsection{Quatitative Results} 

\noindent \textbf{Evaluates the Relevance of Video and Generated Caption} 
To ensure that the descriptions generated by the model are relevant to the video content, Table~\ref{tb:relevance_quality} presents the correlation scores between the input video and the generated descriptions. 
We adopted the method from Video-ChatGPT~\cite{0001RKK24}, using GPT-4o mini to rate the correlation between human annotations and the model-generated descriptions, with scores ranging from 1 to 5. 
Detailed scoring criteria and prompt templates are provided in the appendix. 
The experimental results indicate that our model outperforms others in terms of correlation scores across all video duration intervals. 
Specifically, compared to the baseline model MiniCPMV2.6-8B, our method achieved a 0.41 improvement in correlation scores. 
Even when compared to larger models such as Qwen2-VL-72B and GPT-4o, our model still achieved the best performance. 
This suggests that our model is more accurate and comprehensive in describing the input video content compared to others. 
Notably, MiniCPMV2.6-8B even outperformed Qwen2-VL-72b, which may be due to its training on image datasets. 
We found that, compared to current video captioning datasets, image captioning datasets exhibit longer caption lengths and higher quality.
\begin{figure*}[!t]
  \centering
  \includegraphics[width=0.94\linewidth]{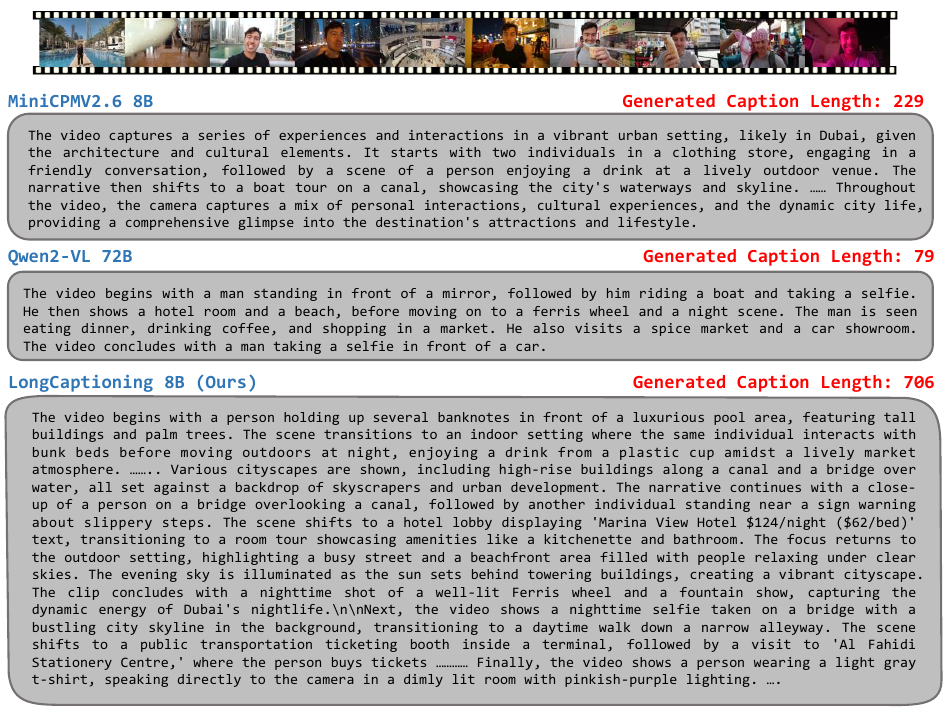}
  \caption{Qualitative comparison of different methods. 
  The results indicate that our proposed method is capable of generating more detailed and comprehensive long-captions compared to other approaches.}
  \label{fig:case}
\end{figure*}

\noindent \textbf{Evaluates the Output Length} 
Table~\ref{tb:length_quality} presents the results of the output length score of the models. 
We report the output length and quality score for videos of different duration intervals. 
For the output length score $S_l$, we use human annotations as the ground truth (assuming the human annotations contain all the content of the test videos) and calculate the difference between the model output length and the human annotation length using Equation~\ref{eq:sl}. 
A higher $S_l$ indicates that the length of the captions generated by the model more closely matches the length of the human annotations. 
Our findings are as follows: 
1) Open-source LMMs generally perform poorly, even those with 72 billion parameters, as they struggle to generate complete captions beyond 300 words. 
2) The performance of MiniCPMV2.6-8B even surpasses that of Qwen2-VL 72B, reflecting that the output length of LMMs is not necessarily correlated with model size. 
3) Notably, compared to other duration intervals, all models show an improvement in output length scores in the [1200s, 1800s) interval. 
After analyzing the generated outputs, we found that some models exhibit repetitive output issues when the input frame is longer, leading to an increase in output length. 
Due to the repetition problem, their quality scores are lower. 
4) By fine-tuning the model on LongCaption-10K, our model achieved optimal performance, even surpassing larger proprietary models like GPT-4o.

Figure~\ref{fig:length_compare} presents the output length statistics for various models, with anomalous outputs like infinitely looping captions manually removed. 
Open-source models, such as MiniCPMV2.6-8B and Qwen2-VL 72B, generated captions primarily between 150 and 300 words. 
The proprietary GPT-4o model produced captions mostly in the 300 to 600-word range. In contrast, our model generated outputs between 300 and 700 words, with a maximum length reaching approximately 1,200 words. 
This demonstrates our model's ability to produce longer, more detailed captions compared to both open-source and proprietary models.

\noindent \textbf{Evaluates the Output Quality} 
Following~\cite{bai2024longwriter}, we further report the output quality of the models in Table~\ref{tb:length_quality}. 
We use GPT-4o mini as the judge to evaluate the output quality across 6 dimensions: Relevance, Accuracy, Coherence, Clarity, Breadth and Depth, and Readability (with each dimension scored on a scale of 1 to 5, and detailed prompt templates provided in the appendix). The average score across these six dimensions is calculated to obtain the overall quality score $S_q$. 
The final results are multiplied by 20 and presented as percentages. 
Our findings are as follows:
1) Larger models like PLLaVA 32B and Qwen2-VL 72B significantly outperform their smaller counterparts (PLLaVA 7B and Qwen2-VL 7B), indicating that increasing model size improves caption quality. 
2) The proprietary model GPT-4o significantly outperforms open-source models. 
Our model achieves the best performance in sentence quality while maintaining output length.


\subsubsection{Ablation Studies}
We explore the effects of removing the visual context window extension and the LongCaption-10K dataset on the LongCaptioning-8B model. 
As shown in Table~\ref{tb:longcaptioning_ablation_relevance} and Table~\ref{tb:longcaptioning_ablation_length}, we find that: 
1) Removing the visual context window extension significantly reduces both output length and quality, with the length score $S_l$ dropping by 8.6, the quality score $S_q$ dropping by 2.1, and the relevance score $S_r$ dropping by 0.24. 
This is because the visual context window extension allows the model to input more sampled frames during the inference process. 
2) Removing the clip-level captioning step during the synthesis of LongCaption-10K leads to a drop in all metrics, as relying solely on frame-level captions fails to capture dynamic changes and action continuity, resulting in less coherent captions. 
3) Similarly, removing the frame-level captioning step further degrades performance, as using only clip-level captions during LongCaption-10K synthesis results in shorter captions. 
4) Without the LongCaption-10K dataset ($i.e.$, using only the MiniCPMV2.6-8B backbone), the model struggles to generate long captions, leading to a sharp decline in output length. Additionally, without both the visual context window extension and the LongCaption-10K dataset, the relevance between generated captions and video content significantly deteriorates.

\subsubsection{The Results of VideoMME}
To further evaluate the performance of LongCaptioning-8B in video understanding, we conducted experiments on the widely used video understanding benchmark, VideoMME~\cite{fu2024video}. 
Table~\ref{tab:video-mme} presents the results of the VideoMME benchmark. 
Compared to the baseline model MiniCPMV2.6-8B, our method consistently demonstrates improvements across all video lengths, including short, medium, and long videos. 
Notably, for long videos, the accuracy increased by 1.6$\%$. 
This phenomenon indicates that incorporating LongCaption-10K during training significantly enhances the model’s ability to understand long videos, providing a foundation for long video captioning.

\subsubsection{The Visualization of LongCaption-10K and LongCaption-Bench}
Figures 7 and 8 display the top 50 word frequency statistics for LongCaption-10K and LongCaption-Bench, respectively. Notably, we excluded non-informative stop words (e.g., "a", "an", "the", "and", "of") to better reveal the primary themes in long-form descriptions. 
A significant divergence in high-frequency terms is clearly observed between LongCaption-10K and LongCaption-Bench, demonstrating notable domain discrepancy between the two datasets. 
This inherent distributional difference effectively mitigates potential performance inflation caused by model overfitting to specific domains during evaluation.

\subsubsection{The Visualization of Long Video Captioning}
Figure~\ref{fig:case} illustrates the qualitative results of video captions generated by the comparison models (MiniCPMV2.6-8B and Qwen2-VL 72B) and our model (LongCaptioning 8B). 
For all models, we uniformly sample 128 frames from the video as input. 
It is evident that LongCaptioning 8B produces outputs with significantly greater length compared to the other models while maintaining the relevance of the generated content to the video. 
In contrast to MiniCPMV2.6-8B and Qwen2-VL 72B, LongCaptioning 8B generates more detailed captions for the input video.


\section{Conclusions}
\label{sec:conclusion}
In this paper, we explore for the first time the long-caption generation capabilities of LMMs and find that current open-source models struggle to produce captions of around 300 words. 
Through controlled experiments, we identify the scarcity of long-caption samples as the primary factor affecting the length of LMM outputs. 
To obtain long-caption samples at a lower cost, we propose LongCaption-Agent, a framework synthesizing long-caption data by aggregating multi-level descriptions. 
Based on this framework, we constructed the LongCaption-10K. 
Additionally, we deployed LongCaption-Bench to reliably evaluate the model's long-caption generation capabilities. 
Using the LongCaption-10K, we successfully extended the caption generation capacity of current LMMs to over 1,000 words, while maintaining high output quality. 
We hope this work will advance research on long-caption generation in LMMs and provide valuable insights for the design of future LMMs.

\bibliography{ms}
\clearpage
\appendix

\section{Training Details}
\begin{table}
    \centering
    \caption{Training Parameters}
    \label{tab:training_parameters}
    \begin{tabular}{lc}
        \toprule
        \textbf{Parameter} & \textbf{Value} \\
        \midrule
        SFT Type & LoRA \\
        $\text{LoRA}_r$ & 8 \\
        $\text{LoRA}_\alpha$ & 32 \\
        Freeze vit & true \\ 
        Batch Size & 1 \\
        Epochs & 1 \\
        Learning Rate & 5e-5 \\
        Gradient accumulation steps & 4 \\
        Dataloader num workers & 1 \\
        Flash attn & true \\
        Deepspeed & zero3-offload \\
        \bottomrule
    \end{tabular}
\end{table}
Table~\ref{tab:training_parameters} reports the details of the training process. 
We conduct training based on the latest open-source models, namely MiniCPMV2.6-8B~\cite{yao2024minicpm}. 
This model has only undergone pretraining and supervised fine-tuning (SFT) on image-text datasets. 
This allows us to eliminate the interference of other video datasets when analyzing the experimental results. 
We employ the LoRA strategy for supervised fine-tuning. 
Specifically, we fine-tune only a small subset of parameters in the LLM, where the $\text{LoRA}_r$ is set to 8 and the $\text{LoRA}_\alpha$ is set to 32. 
Moreover, to further extend the effective context window during inference and allow the input of longer frame sequences, we introduced a visual context window extension technique~\cite{wei2024visual} during training. 
This technique scales the rotational frequency of the position encoding embedding for visual tokens, thereby extending the model's context window. 
The model was trained on a 2 A800-80G GPUs using DeepSpeed + ZeRO3 + CPU offloading~\cite{AminabadiRALLZRSZRH22}. 
We used a batch size of 1, a learning rate of 5e-5, and trained for 1 epoch.

\section{Prompts}
\label{sec:prompt}
\subsection{Test Prompt}
In Figure~\ref{fig:exp1}, all models use the same prompt. 
\begin{tcolorbox}[title = { }]
  \scriptsize
    \noindent
    \emph{Task}:
  
    You are an expert in understanding scene transitions based on visual features in a video. You are requested to create the descriptions for the current video sent to you,  which includes multiple sequential frames.
  
    \emph{Guidelines For Video Description}:
    
    - Analyze the narrative progression implied by the sequence of frames, interpreting the sequence as a whole.
    - Note that since these frames are extracted from a video, adjacent frames may show minimal differences. These should not be interpreted as special effects in the video.
    - If text appears in the frames, you must describe the text in its original language and provide an English translation in parentheses. For example: book. Additionally, explain the meaning of the text within its context.
    - When referring to people, use their characteristics, such as clothing, to distinguish different people.
    - **IMPORTANT** Please provide as many details as possible in your description, including colors, shapes, and textures of objects, actions and characteristics of humans, as well as scenes and backgrounds. 
   
    \emph{Output Format}:
  
    Your response should look like this: {"Video Level Description": "The video begins with..., progresses by..., and concludes with..."}

    Please give me the description of the current video.
  
  \end{tcolorbox}


  \subsection{Quality Score}
The prompt used when evaluating the quality score is as follows.
\begin{tcolorbox}[title = { }]
  \scriptsize
    \noindent
    You are an expert in evaluating text quality. Please evaluate the quality of an AI assistant's response to a user's writing request. Be as strict as possible.

    You need to evaluate across the following six dimensions, with scores ranging from 1 to 5. The scoring criteria from 5 to 1 for each dimension are as follows:

    1. Relevance: From content highly relevant and fully applicable to the user's request to completely irrelevant or inapplicable.

    2. Accuracy: From content completely accurate with no factual errors or misleading information to content with numerous errors and highly misleading.

    3. Coherence: From clear structure with smooth logical connections to disorganized structure with no coherence.

    4. Clarity: From clear language, rich in detail, and easy to understand to confusing expression with minimal details.

    5. Breadth and Depth: From both broad and deep content with a lot of information to seriously lacking breadth and depth with minimal information.

    6. Reading Experience: From excellent reading experience, engaging and easy to understand content to very poor reading experience, boring and hard to understand content.
    
    Please evaluate the quality of the following response to a user's request according to the above requirements.

    $\langle User\quad Request \rangle$ : 

    $\langle /Response \rangle$ : 

    '''Please evaluate the quality of the response. You must first provide a brief analysis of its quality, then give a comprehensive analysis with scores for each dimension. The output must strictly follow the JSON format: {"Analysis": ..., "Relevance": ..., "Accuracy": ..., "Coherence": ..., "Clarity": ..., "Breadth and Depth": ..., "Reading Experience": ...}. You do not need to consider whether the response meets the user's length requirements in your evaluation. Ensure that only one integer between 1 and 5 is output for each dimension score.'''

  \end{tcolorbox}

  \subsection{Video-Caption Relevance Score}
The prompt used when evaluating the Video-Caption Relevance Score is as follows.
\begin{tcolorbox}[title = { }]
  \scriptsize
    \noindent
    You are an intelligent chatbot designed for evaluating the detail orientation of generative outputs for video-based question-answer pairs. "
    "Your task is to compare the predicted answer with the correct answer and determine its level of detail, considering both completeness and specificity. Here's how you can accomplish the task:"
    
    "------"

    \emph{INSTRUCTIONS}: 

    "- Check if the predicted answer covers all major points from the video. The response should not leave out any key aspects."
    "- Evaluate whether the predicted answer includes specific details rather than just generic points. It should provide comprehensive information that is tied to specific elements of the video."
    "- Consider synonyms or paraphrases as valid matches."
    "- Provide a single evaluation score that reflects the level of detail orientation of the prediction, considering both completeness and specificity."

    Please evaluate the following video-based question-answer pair:

    $\langle User\quad Request \rangle$ : 

    $\langle Correct\quad Answer \rangle$ : 

    $\langle Predicted\quad Answer \rangle$ : 

    "Provide your evaluation only as a detail orientation score where the detail orientation score is an integer value between 0 and 5, with 5 indicating the highest level of detail orientation. "
    "Please generate the response in the form of a Python dictionary string with keys 'score', where its value is the detail orientation score in INTEGER, not STRING."
    "DO NOT PROVIDE ANY OTHER OUTPUT TEXT OR EXPLANATION. Only provide the Python dictionary string. "
    "For example, your response should look like this: {'score': 4.8}."
  
  \end{tcolorbox}

  \subsection{Frame-Level Captioning} 
We present our prompt in use: 
\begin{tcolorbox}[title = { }]
  \scriptsize
  \noindent
  \emph{Task}:

  You are an expert in understanding the visual details of individual frames within a video. You are requested to create detailed descriptions for each video frame sent to you. Your task is to describe the frame's content with high precision, focusing only on the elements visible in that exact frame. Do not infer or speculate about actions or events not explicitly visible in the frame.

  \emph{Guidelines For Frame Description}:
  
  - Describe only what is visible in the frame: Focus on the exact visual details, without making assumptions about what happens before or after.
  - Avoid narrative progression: Unlike a clip description, there is no need to interpret or connect this frame with others. Only describe the current frame.
  - Be specific and exhaustive: Include as many details as possible, such as:
    - Objects: Colors, shapes, textures, positions, and relationships between objects.
    - People: Clothing, facial expressions, posture, gestures, and any visible features (e.g., hair color, accessories).
    - Background: Environmental details, lighting, shadows, and any visible text (with translations if necessary).
    - Text in the frame: If text appears, provide its original language and an English translation in parentheses.
  - No additional reasoning: Do not infer motivations, future actions, or unseen parts of the scene.
 
  \emph{Output Format}:

  Your response should look like this: {"Frame Level Description": "The frame shows..."}

\end{tcolorbox}

\subsection{Clip-Level Captioning} 
Here is the prompt we use: 
\begin{tcolorbox}[title = { }]
  \scriptsize
  \noindent
  \emph{Task}:

  You are an expert in understanding scene transitions based on visual features in a video. There is a video including multiple sequential clips (clip-1,clip-2,...). Given the description for these clips (clip-1,clip-2,...,) as the context, you are requested to create the descriptions for the current clip sent to you,  which includes multiple sequential frames.

  \emph{Guidelines For Clip Description}:
  
  - Your description should see the description of previous clips as context.
  - Analyze the narrative progression implied by the sequence of frames, interpreting the sequence as a whole.
  - Note that since these frames are extracted from a clip, adjacent frames may show minimal differences. These should not be interpreted as special effects in the clip.
  - Note that some objects and scenes shown in the previous clips might not shown in the current clip. Be carefully do not assume the same object and scenes shown in every clips.
  - If text appears in the frames, you must describe the text in its original language and provide an English translation in parentheses. For example: book. Additionally, explain the meaning of the text within its context.
  - When referring to people, use their characteristics, such as clothing, to distinguish different people.
  - **IMPORTANT** Please provide as many details as possible in your description, including colors, shapes, and textures of objects, actions and characteristics of humans, as well as scenes and backgrounds. 
 
  \emph{Output Format}:

  Your response should look like this: {"Clip Level Description": "The clip begins with..., progresses by..., and concludes with..."}

  Description of Previous Clips: \{\emph{t-1\_step\_clip\_description}\}
\end{tcolorbox}

\subsection{Video-Level Captioning} 
\noindent \textbf{Video-Level Captioning} 
The prompt used is: 
\begin{tcolorbox}[title = { }]
  \scriptsize
  \noindent
  \emph{Task}:

  You are an expert at understanding frame-level and clip-level descriptions in a video that includes \{\emph{num\_frame}\} frames and \{\emph{num\_clip}\} clips. You are requested to create a video description by summarizing these frame-level and clip-level descriptions chronologically.

  \emph{Guidelines For Video Description}:
  
  - Since the frame-level and clip-level descriptions are provided in chronological order, ensure that the video description is coherent and follows the same sequence. Avoid referring to the first or final frame of each clip as the first or final frame of the entire video.
  - Include any text that appears in the clip, provide its English translation in parentheses, and explain the significance of each text within its context.
  - The tone of the video description should be as if you are describing a video directly instead of summarizing the information from several clip descriptions. Therefore, avoid phrases found in the referred clip descriptions such as "The clip begins...", "As the clip progresses...", "The clip concludes", "The final/first frame", "The second clip begins with", "The final frames of this segment", etc
  - **IMPORTANT** Include all details from the given clip descriptions in the video description. Try to understand of the theme of the video and provide a coherent narrative that connects all the clips together.
 
  \emph{Output Format}:

  1. Your output should be formed in a JSON file.
  2. Only provide the Python dictionary string.
  3. You can use various descriptive sentence structures to outline the narrative progression. One example is: \{\}
  Your response should look like this: \{\{"Video Level Description": "YOUR DESCRIPTION HERE."\}\}

  Frame-level Description (sorted in chronological order by number): \{\emph{t\_step\_frame\_descriptions}\}

  Clip-level Description (sorted in chronological order by number): \{\emph{t\_step\_clip\_description}\}

  Please give me the description of the video given the frame-level and clip-level descriptions.
  
\end{tcolorbox}

\begin{figure*}[!h]
    \centering
    \includegraphics[width=0.99\linewidth]{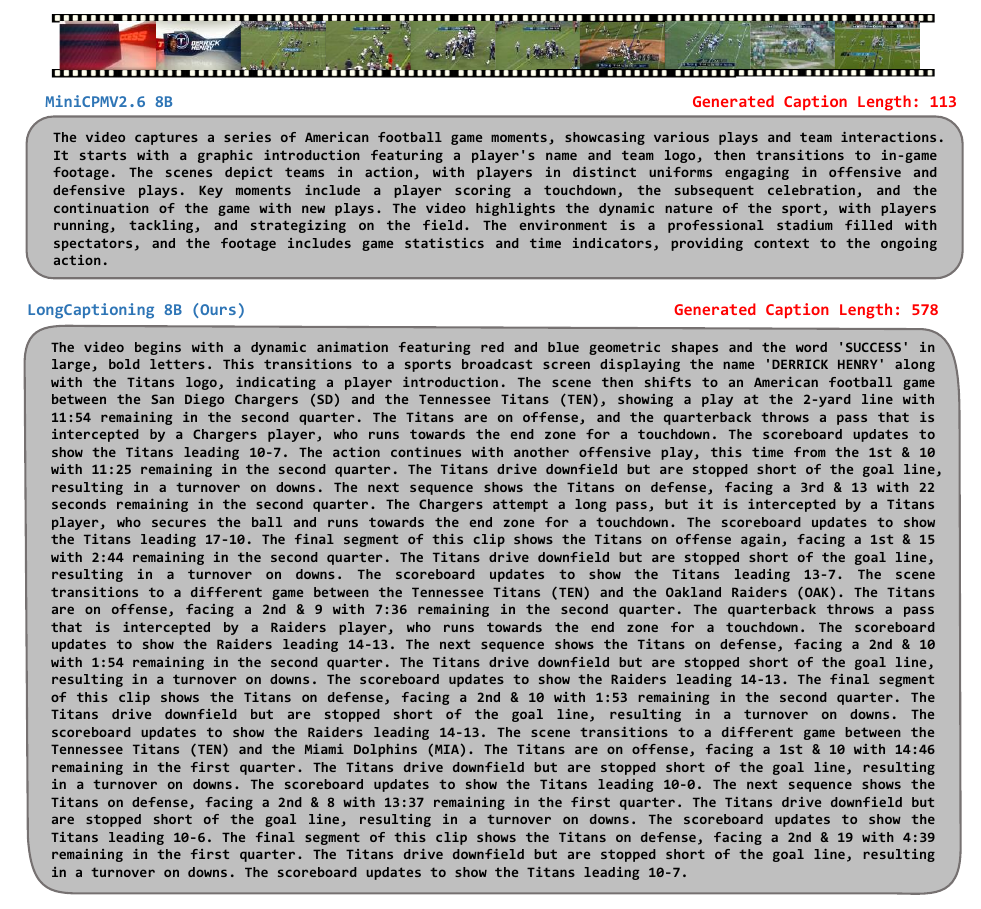}
    \caption{Qualitative comparison of different methods. }
    \label{fig:case_app}
  \end{figure*}

  \begin{figure*}[!h]
    \centering
    \includegraphics[width=0.99\linewidth]{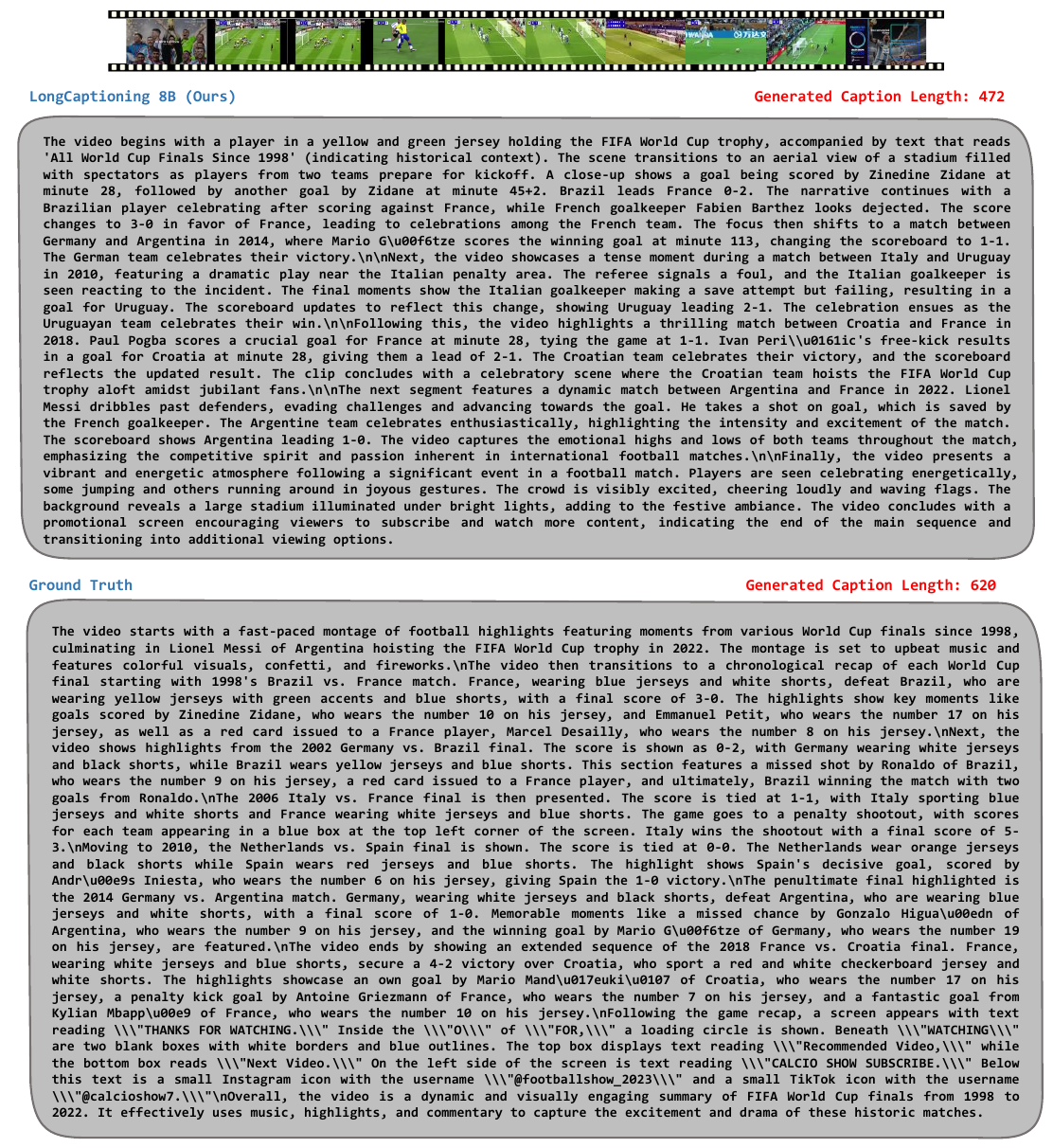}
    \caption{Qualitative comparison of LongCaptioning-8B and Ground Truth. }
    \label{fig:case_app_2}
  \end{figure*}

\end{document}